\documentclass{article}

\usepackage{arxiv}
\usepackage[utf8]{inputenc} 
\usepackage[T1]{fontenc}    
\usepackage{hyperref}       
\usepackage{url}            
\usepackage{booktabs}       
\usepackage{amsfonts}       
\usepackage{nicefrac}       
\usepackage{microtype}      
\usepackage{lipsum}
\usepackage{graphicx}
\graphicspath{ {./images/} }
\usepackage{threeparttable}
\usepackage{bbold}
\usepackage{bm,nicefrac}
\usepackage{float}
\usepackage{tabularray}
\usepackage{adjustbox}
\usepackage{algorithm}
\usepackage[algo2e]{algorithm2e} 

\usepackage{algorithmic}
\usepackage[numbers]{natbib}

\usepackage{graphicx}
\usepackage{subfigure}
\usepackage{booktabs} 

\usepackage[utf8]{inputenc} 
\usepackage[T1]{fontenc}    
\usepackage{hyperref}       
\usepackage{url}            
\usepackage{booktabs}       
\usepackage{amsfonts}       
\usepackage{nicefrac}       
\usepackage{microtype}      
\usepackage{xcolor}         
\usepackage{amsmath}
\usepackage{amssymb}
\usepackage{mathtools}
\usepackage{amsthm}

\usepackage[capitalize,noabbrev]{cleveref}
\usepackage{float}

\usepackage{amssymb}
\usepackage{mathtools}
\DeclarePairedDelimiter{\ceil}{\lceil}{\rceil}

\theoremstyle{plain}
\newtheorem{theorem}{Theorem}

\newtheorem{lemma}{Lemma}

\theoremstyle{definition}
\newtheorem{definition}{Definition}
\newtheorem{claim}{Claim}

\theoremstyle{remark}

\newcommand{\ra}{\rightarrow}

\newcommand{\prob}[1]{\mathbb{P}\left(#1\right)}

\newcommand{\I}{\mathcal{I}}
\newcommand{\A}{\mathcal{A}}

\newcommand{\ignore}[1]{}
\allowdisplaybreaks[4]

\title{Representative Arm Identification: A fixed confidence approach to identify cluster representatives}

\author{
  Sarvesh Gharat \thanks{sarveshgharat19@gmail.com}\\
  Centre for Machine Intelligence and Data Science\\
  IIT Bombay\\
   \And
  Aniket Yadav \\
  Centre for Machine Intelligence and Data Science\\
  IIT Bombay\\
  \And
  Nikhil Karamchandani \\
  Department of Electrical Engineering\\
  IIT Bombay\\
  \And
  Jayakrishnan Nair \\
  Department of Electrical Engineering\\
  IIT Bombay\\
}

\begin{document}
\maketitle
\begin{abstract}

We study the representative arm identification (RAI) problem in the multi-armed bandits (MAB) framework, wherein we have a collection of arms, each associated with an unknown reward distribution. An underlying instance is defined by a partitioning of the arms into clusters of predefined sizes, such that for any $j > i$, all arms in cluster $i$ have a larger mean reward than those in cluster $j$. The goal in RAI is to reliably identify a certain prespecified number of arms from each cluster, while using as few arm pulls as possible. The RAI problem covers as special cases several well-studied MAB problems such as identifying the best arm or any $M$ out of the top $K$, as well as both full and coarse ranking. We start by providing an instance-dependent lower bound on the sample complexity of any feasible algorithm for this setting. We then propose two algorithms, based on the idea of confidence intervals, and provide high probability upper bounds on their sample complexity, which orderwise match the lower bound. Finally, we do an empirical comparison of both algorithms along with an LUCB-type alternative on both synthetic and real-world datasets, and demonstrate the superior performance of our proposed schemes in most cases. 
\end{abstract}

\section{Introduction}
\label{sec: Introduction}
The stochastic multi-armed bandit (MAB) problem \cite{lattimore2020bandit} is a  widely studied online decision-making framework, which consists of $K$ arms each associated with an a priori unknown reward distribution. Each pull of an arm results in a random reward, generated i.i.d. from the associated distribution. At each round, the learner can decide which arm to pull based on the entire history of pulls and rewards. The MAB framework has found application in a wide variety of domains ranging from A/B testing \cite{kaufmann2014complexity} and online advertising \cite{geng2021comparison} to network routing \cite{talebi2017stochastic}, clinical testing \cite{villar2015multi}, and hyperparameter optimization \cite{li2018hyperband} in machine learning. 

There are several objectives that a learner might be interested in while interacting with the MAB. For example, one widely studied goal is to maximize the expected cumulative reward accrued by the learner over a certain time horizon, or equivalently to minimize the \textit{regret} with respect to an oracle which knows the arm reward distributions beforehand. Several regret minimization algorithms have been proposed in the literature \cite{agrawal2017near,bubeck2012regret}, and they are typically based on the idea of balancing \textit{exploration} (trying different arms to reduce uncertainty about their mean rewards) and \textit{exploitation} (pulling the arms known to have high rewards). 

Another popular learning objective is identifying the best arm in terms of the mean reward \cite{kaufmann2016complexity, jamieson2014best}. The problem has been studied in both the \textit{fixed confidence} setting \cite{jamieson2014best}, where the goal is to find the best arm using the minimum number of pulls while guaranteeing a certain pre-specified error probability $\delta$; and the \textit{fixed budget} setting \cite{audibert2010best} where the total number of pulls is fixed beforehand and the aim is to minimize the probability of error. In our work, we will focus on the fixed confidence setting. Several such \textit{pure exploration} problems beyond best arm identification have been studied in the literature, including identifying the top $K$ arms \cite{kalyanakrishnan2012pac} or the arms with mean rewards above a given threshold \cite{locatelli2016optimal}. These problems can often require a very large number of samples to solve, which makes them impractical for many applications of interest. One way of rectifying this shortcoming is to relax the objective, for example to identifying an $\epsilon$-best arm \cite{karnin2013almost} whose mean reward is within some small $\epsilon$ gap to the best arm, or any $M$ out of the top $K$ arms \cite{chaudhuri2017pac, chaudhuri2019pac}, or any `good' arm whose mean reward is above a threshold \cite{degenne2019pure}. All the above relaxations can be treated as MAB problems with \textit{multiple correct answers}, which has been studied recently in \cite{degenne2019pure} wherein a general lower bound on the sample complexity and an asymptotically optimal (as error probability $\delta \rightarrow 0$) algorithm are provided.  

Along similar lines, in this work, we propose the Representative Arm Identification (RAI) problem for which an underlying MAB instance is defined by a partitioning of the arms into clusters of predefined sizes, such that for any $j > i$, all arms in cluster $i$ have a larger mean reward than those in cluster $j$. The goal in RAI is to reliably identify a certain prespecified number of arms from each cluster, while using as few arm pulls as possible. The RAI problem naturally arises in several real-world applications. For example, consider a crowdsourcing platform which engages a collection of workers with apriori unknown skill levels. Say, the platform is hired to solve a large task, which can be broken into multiple sub-tasks of differing complexity. Then the platform might want to hire some workers amongst the best few to tackle the hard subtasks, some with around average skill level to deal with subtasks of medium hardness, and then a few with the lowest skill level to handle the easiest subtasks. Another application of the RAI problem is in online recommendation systems, for example a content creator might be interested in knowing a few movies amongst the best rated, and also some amongst the worst rated on the platform.

Beyond several natural applications, the RAI problem is also interesting from a theoretical viewpoint since it  covers as special cases several well-studied MAB problems such as best arm identification \cite{jamieson2014best} or 
identifying any $M$ out of the top $K$ arms \cite{chaudhuri2019pac}, as well as both full and coarse ranking \cite{karpov2020batched}, and is able to provide a unifying viewpoint for all these problems; see Table~\ref{tab: standard examples} for a list of such problems. We are able to provide a lower bound on the sample complexity of the general RAI problem, in terms of a quantity we call the `bottleneck gap' which depends on the underlying instance and the arms requirements, and encodes the hardness of the problem. We also present two algorithms, based on the idea of confidence intervals,   for reliably solving any RAI problem and provide high probability upper bounds on the sample complexity of these schemes. Finally, we conduct an empirical comparison of both our algorithms along with a LUCB-type baseline on both synthetic and real-world datasets, and demonstrate the superior performance of our proposed schemes in most cases.  

\section{Problem Formulation}
\label{sec: Problem Formulation}

We consider a stochastic multi-armed bandit with a collection $\mathcal{N}$ of $N=| \mathcal{N}|$ arms, each associated with a $\frac{1}{2}$-subGaussian reward distribution\footnote{A random variable $X$ is $\sigma$-subGaussian if, for any $t > 0$, $\mathbb{P} \left(|X - \mathbb{E}[X]|>t\right) \leq 2\exp \left(-t^2/2\sigma^2 \right).$}, which is a priori  unknown to the learner.

To define the representative arm identification (RAI) problem, we sort the arms in decreasing order of their mean rewards and then partition them into $m$ clusters of predefined sizes given by $c = (c_1, c_2, \cdots, c_m)$, such that for any $j > i$, all arms in cluster $i$ have a larger mean reward than those in cluster $j$. Clearly, $\sum_{i=1}^m c_i = N.$ 

\begin{figure}[H]
    \hspace*{-1cm}
    \includegraphics[scale = 0.11]{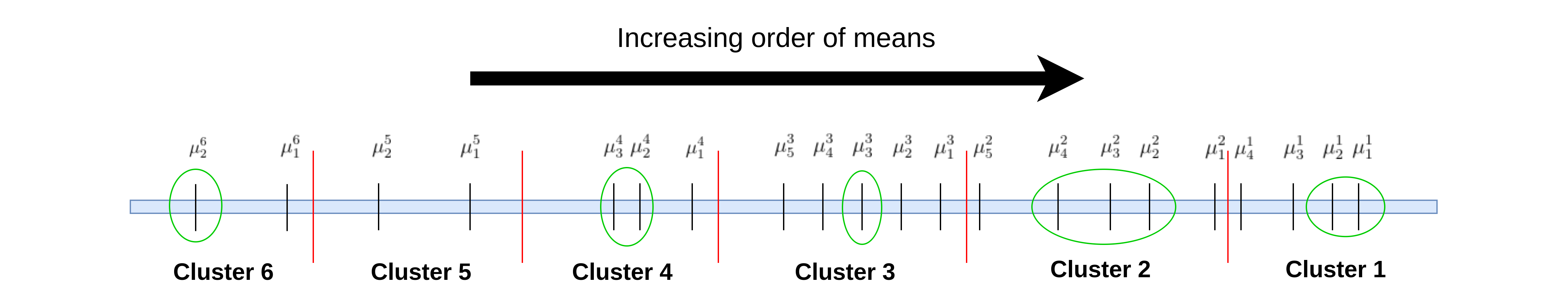}
    \caption{An RAI problem instance with $m=6$ clusters, $c = (4,5,5,3,2,2),$ and $r = (2,3,1,2,0,1).$ The circled arms illustrate one of the correct outputs for this problem.}
    \label{fig: exampleinstance}
\end{figure}

We label the arms as follows: for each $i \in [m] := \{1,2,\ldots,m\}$ and $j \in [c_i] := \{1,2,\ldots,c_i\}$, we have an associated reward distribution $\Pi_{j}^{i}$ with the $j$-th arm in cluster $i$, so that each pull of the arm results in an i.i.d. reward sample from $\Pi_{j}^{i}$. The corresponding mean reward is denoted by $\mu_j^i$, and we have that $\mu_{j_1}^i \ge \mu_{j_2}^i$ for any $j_1 \ge j_2$. See Figure ~\ref{fig: exampleinstance} for an illustration. We assume this partition of arms into clusters is uniquely defined, i.e., if $i_1 \neq i_2, \; \nexists \; j_1, j_2 \text{ such that } \mu_{j_1}^{i_1} = \mu_{j_2}^{i_2}$. 

Finally, we have a prespecified vector $r = (r_1, r_2, \ldots, r_m)$ such that $0 \leq r_i \le c_i$ for all $i$, and where $r_i$ denotes the number of `representative' arms belonging to cluster $i$ that the learner needs to identify. As mentioned before, 
the RAI problem covers a wide range of very well-studied MAB problems, ranging from the best arm identification problem, where  $c = (1, N-1), r = (1,0),$ to full ranking, where $c = (1, 1, \cdots, 1), r = (1, 1, \cdots,1).$ Table \ref{tab: standard examples} lists several classical learning tasks from the literature that are special cases of the RAI problem. An important aspect of the RAI problem is that there can be \emph{multiple correct answers}; specifically, this is the case if $0 < r_i < c_i$ for some~$i \in [m].$ 

To summarize, an instance of the RAI problem is defined by $\I = (c,r,\Pi),$ where $\Pi = (\Pi^i_j, i \in [m],j\in [c_i]).$ To perform the RAI task, the learner can use an online algorithm, say $\mathcal{A}$, which at each time can either choose an arm to sample based on past observations; or decide to stop and output an estimated collection of representative arms from each cluster, given by
$(O_i, i \in [m]);$ here, $O_i$ denotes the set of representatives from cluster~$i,$ with $|O_i| = r_i.$
Given a prespecified error threshold $\delta \in (0,1),$ we say that the algorithm $\mathcal{A}$ is $\delta$-probably correct ($\delta$-PC) if, for any underlying problem instance $\mathcal{I}$, the probability that algorithm output is incorrect is at most~$\delta.$ 
More formally, denoting the (random) stopping time of the algorithm~$\mathcal{A}$ by $T^{\mathcal{I}}_{\delta}(\mathcal{A}),$ $\A$ is $\delta$-PC if, for any instance~$\I,$
$$\prob{T^{\mathcal{I}}_{\delta}(\mathcal{A}) < \infty,\ \exists \ i \in [m] \mbox{ and } j \in O_i \mbox{ s.t. arm~} j \mbox{ is not in cluster } i} \leq \delta.$$

The performance of a $\delta$-PC algorithm $\mathcal{A}$ is captured via its sample complexity $T^{\mathcal{I}}_{\delta}(\mathcal{A})$, i.e., the number of arm pulls it makes before stopping. Our goal in this paper is to design $\delta$-PC schemes for the RAI problem, whose sample complexity $T^{\mathcal{I}}_{\delta}(\mathcal{A})$ is as small as possible. Note that $T^{\mathcal{I}}_{\delta}(\mathcal{A})$ is itself a random quantity, and our results will be in terms of expectation or high probability bounds.

\begin{table}[H]
\caption{MAB problems from the literature that are special cases of the RAI problem}
\label{tab: standard examples}
\begin{center}
\begin{tabular}{| l | c | c | c | c |} 
 \hline
 Task & Number of Clusters & $c$ & $r$ \\ 
 \hline\hline
 Best Arm Identification \citep{kaufmann2016complexity} & $2$ & $(1, N-1)$ & $(1, 0)$ \\ 
 \hline
 One out of Top $K$ \citep{chaudhuri2017pac} & $2$ & $(K, N-K)$ & $(1, 0)$ \\
 \hline
 Top $K$ arm identification \citep{kalyanakrishnan2012pac} & $2$ & $(K, N-K)$ & $(K, 0)$ \\
 \hline
 $M$ out of Top $K$ \citep{chaudhuri2019pac} & $2$ & $(K, N-K)$ & $(M, 0)$ \\
 \hline
 Coarse Ranking \citep{karpov2020batched} & $m$ & $(c_1, c_2, \cdots, c_m)$ & $(c_1, c_2, \cdots, c_m)$ \\ 
 \hline
 Full Ranking \citep{karpov2020batched} & $N$ & $(1, 1, \cdots, 1)$ & $(1, 1, \cdots, 1)$ \\
 \hline
\end{tabular}
\end{center}
\end{table}

\section{Lower Bound}
\label{sec: Lower Bound}

In this section, we derive a lower bound on the sample complexity of any $\delta$-PC algorithm for the RAI problem. To state the result, we first need a few definitions. 

\begin{definition}
\label{def: arm gap}
(Arm Gap): Consider an instance $\mathcal{I} = (c,r,\Pi);$ recall that $\mu_j^i$ denotes the mean reward of the $j^{\text{th}}$ arm in cluster $i$. The \emph{arm gap} $\Delta_j^i$ for that arm is defined as
$$\Delta_j^i := \min\{ \mu_{c_{i-1}}^{i-1} - \mu_j^i, \mu_j^i - \mu_1^{i+1}\}, \quad \forall j \in c_i, i \in [m].$$ Here, for notational simplicity, we have assumed two dummy clusters, $0$ and $m+1,$ having one arm each, such that $\mu^0_1 = \infty,$ $\mu^{m+1}_1 = -\infty.$
\end{definition}
In words, $\Delta_j^i$ represents the the gap (in mean reward) between the $j^{\text{th}}$ arm in cluster $i$ and the `nearest' arm in a neighboring cluster.

Next, we define the bottleneck gap associated with an instance. 

\begin{definition} 
\label{def: optimal gap}
(Bottleneck Gap): Consider an instance $\mathcal{I} = (c,r,\Pi).$ For each cluster $i \in [m]$, let $\Delta^i_{\mathcal{I}}$ denote the $r_i$-\text{th} largest arm gap amongst its arms. By convention, $\Delta^i_{\mathcal{I}} = \infty$ if $r_i = 0.$ The \emph{bottleneck gap} $\Delta_{\mathcal{I}}$ associated with the instance~$\I$ is defined as  
$$\Delta_{\mathcal{I}} := \min \{\Delta^1_{\mathcal{I}}, \Delta^2_{\mathcal{I}}, \cdots , \Delta^m_{\mathcal{I}}\} .$$
\end{definition}

Intuitively, $\Delta^i_{\mathcal{I}}$ captures the complexity associated with the sub-task of identifying~$r_i$ arms from cluster~$i,$ while $\Delta_{\mathcal{I}}$ captures the complexity associated with the complete RAI task; the smaller these gaps, the harder the corresponding task. This is formalized in the following theorem.

\begin{theorem}
\label{theorem: lower bound}
For a given error threshold $\delta \in (0,1),$ in the space of $\frac{1}{2}$-Gaussian instances (i.e., each arm has a Gaussian reward distribution with standard deviation $\sigma = \frac{1}{2}$), any $\delta$-PC algorithm $\mathcal{A}$ for the RAI problem 
satisfies $$\liminf_{\delta \ra 0} \dfrac{E[T^{\mathcal{I}}_{\delta}(\mathcal{A})]}{\log(1/\delta)} \geq \dfrac{1}{2(\Delta_{\mathcal{I}})^2}$$ 
\end{theorem}
The proof of Theorem \ref{theorem: lower bound} is provided in Appendix \ref{Theorem Proofs}. As mentioned before, the RAI problem falls in the class of MAB problems with multiple correct answers and as such, the general lower bound provided in \cite{degenne2019pure} applies to the RAI problem as well. 
While the lower bound in~\cite{degenne2019pure} is in the form of a $\min \min \max$ optimization problem, the result above specializes it to the RAI problem, and also relaxes it to provide a simpler, more interpretable bound in terms of the bottleneck gap of the underlying instance. While the lower bound in Theorem~\ref{theorem: lower bound} is in general looser than the (less explicit) bound that follows from~\cite{degenne2019pure}, it captures the core complexity of the RAI task, and also enables a direct comparison to upper bounds on the sample complexity of our proposed algorithms (see Section~\ref{sec: Algorithms}). 

We conclude this section by specializing the lower bound in Theorem~\ref{theorem: lower bound} to the task of identifying~$M$ out of the top~$K$ arms, where $1 \leq M \leq K < N$ (this special case of the RAI problem was analysed in~\citep{chaudhuri2019pac}).\footnote{Note that this task further subsumes other classical tasks like best arm identification ($M=K=1$), and the identification of one of the top~$K$ arms (analysed in~\citep{chaudhuri2017pac}) as special cases.} Without loss of generality, suppose that the arms are also labelled $1,2,\ldots,N$ such that the corresponding mean rewards satisfy~$\mu_1 \geq \mu_2 \geq \cdots \geq \mu_N.$ Then it is easy to show that the bottleneck gap for this task is given by $\Delta_{\mathcal{I}} = \mu_{M}-\mu_{K+1}.$ To the best of our knowledge, such an explicit, interpretable, instance-dependent complexity characterization is not available in the literature for this task. 

\section{Algorithms}
\label{sec: Algorithms}

This section describes the two algorithms we propose to solve the RAI problem and presents upper bounds on their sample complexity. In spirit, these algorithms are similar to the \textit{successive elimination} style schemes which are widely used for MAB problems~\cite{jamieson2014best}. Under both algorithms, active arms are pulled in a round robin fashion, and suitable confidence intervals are maintained for the mean reward of each active arm. From time to time, arms whose membership in a certain cluster can be inferred based on the computed confidence intervals are `selected' to be part of the algorithm output; these selected arms are then removed from the active set. The two algorithms differ with respect to the scheduling of the membership check---the \emph{Vanilla Round Robin Algorithm} performs this check after each round robin cycle, whereas the \emph{Butterscotch Round Robin Algorithm} performs the membership check only on each halving of the confidence interval widths. 

\noindent {\bf Vanilla Round Robin Algorithm for RAI}

The Vanilla Round Robin Algorithm is stated formally as Algorithm~\ref{algo: vanilla}. This algorithm proceeds in rounds; each round involves the following steps: 
\begin{itemize}
    \item The algorithm samples every arm in its active set $A$ once (line 4); this is the set of arms whose cluster membership is not yet confirmed.
    \item  The arms in~$A$ are then partitioned into tentative clusters of sizes $\tilde{c}_1,\cdots,\tilde{c}_m$ based on their empirical mean rewards (line~6); here, $\tilde{c_i}$ is the (estimated) number of active arms from cluster~$i$ (more formally, $\tilde{c}_i = c_i - |O_i|,$ where $O_i$ denotes the set of representative arms that have been identified by the algorithm as being from cluster~$i$ by that point).
    \item Next, for each arm~$a$ in the active set, a check is made to see whether its cluster assignment, say~$i,$ can be finalised. This entails  checking whether the confidence intervals indicate arm~$a$ as being `better' than $\sum_{j > i} \tilde{c}_i$ active arms, and `worse' than $\sum_{j < i} \tilde{c}_i$ active arms (line~11). If this holds, arm~$a$ is assigned to cluster~$i$ (if additional representatives are required from cluster~$i),$ and it is removed from the active set.
    \item Finally, we merge adjacent clusters whose requirement for  representative arms has been met (lines 19--22). Such a merger relaxes the membership check for the merged cluster (note that the algorithm does not need to output any arms from the merged cluster), hastening the removal of its members from the active set. Empirically, we find that this final step results in significantly fewer arm pulls by the algorithm, as we demonstrate in Section~\ref{sec: numerical}.
\end{itemize}
Of course, the algorithm terminates when the requisite number of representatives have been identified from each cluster.
The following result establishes that the Vanilla Round Robin Algorithm is $\delta$-PC, and also provides an instance-dependent high probability upper bound on the sample complexity of the algorithm.

\begin{algorithm}[t]
\caption{Vanilla Round Robin Algorithm for RAI}
\label{algo: vanilla}
\textbf{Input}: cluster sizes $c = (c_1, c_2, \cdots, c_m)$, required arms $r = (r_1, r_2, \cdots, r_m)$, arm set $\mathcal{N}$, error threshold~$\delta$ \\
\textbf{Output}: $O_1, O_2, \cdots, O_m$
\begin{algorithmic}[1]
\STATE Initialize $R \leftarrow 0, A \leftarrow \mathcal{N},$ and for $i \in \{1, 2, \cdots, m\}$ set $\tilde{c}_i = c_i$ 
\WHILE{$|O_1| \neq r_1 \text{ or } |O_2| \neq r_2 \text{ or } \cdots |O_m| \neq r_m$}
\STATE Increment 
$R$ by $1$
\STATE Sample every arm in $A$ once
\STATE Update empirical mean rewards
\STATE 
Partition~$A$ into clusters $A_1, A_2, \cdots, A_m$ of sizes $\tilde{c}_1, \tilde{c}_2, \cdots, \tilde{c}_m$ respectively, based on the empirical means
\FOR{$i$ in $[m]$}
\FOR{arm $a$ in $A_i$}
\STATE $\mathrm{Better}(a) = \left\{j \in \cup_{j < i} A_j \colon \hat{\mu}_j - \hat{\mu}_a > 2\sqrt{\frac{\ln(\pi^2R^2N/3\delta)}{2R}}  \right\}$ \\
\STATE $\mathrm{Worse}(a) = \left\{j \in \cup_{j > i} A_j \colon \hat{\mu}_a - \hat{\mu}_j > 2\sqrt{\frac{\ln(\pi^2R^2N/3\delta)}{2R}}  \right\}$ \\
\IF {$|\mathrm{Better}(a)| \geq \sum_{j < i} \tilde{c}_j$ and $|\mathrm{Worse}(a)| \geq \sum_{j > i} \tilde{c}_j$}
\IF{$|O_i| < r_i$}
\STATE Add arm~$a$ to $O_i$
\ENDIF
\STATE Remove arm~$a$ from $A$ 
\STATE $\tilde{c}_i \leftarrow \tilde{c}_i - 1$
\ENDIF
\ENDFOR
\IF{$i > 1$ and $|O_{i-1}| = r_{i-1}$ and $|O_{i}| = r_{i}$}
\STATE $\tilde{c}_i \leftarrow \tilde{c}_{i-1} + \tilde{c}_i$
\STATE $\tilde{c}_{i-1} \leftarrow 0$
\ENDIF
\ENDFOR
\ENDWHILE
\end{algorithmic}
\end{algorithm}
\begin{theorem} 
\label{theorem: vanilla}
    The Vanilla Round Robin Algorithm (see Algorithm~\ref{algo: vanilla}) is $\delta$-PC for the RAI problem. With probability at least $1-\delta$, its sample complexity $\mathcal{T}_\delta^{\mathcal{I}}$ satisfies
    $$\mathcal{T}_\delta^{\mathcal{I}} \le  \sum_{i=1}^{m} \sum_{j = 1}^{c_i}\Biggl(\mathbb{1}\{\Delta_j^i \geq \Delta_{\mathcal{I}}\} \dfrac{26}{(\Delta_j^i)^2} \ln\Biggl(\dfrac{16\pi\sqrt{\dfrac{N}{3\delta}}}{ (\Delta_j^i)^2}\Biggr) + \mathbb{1}\{\Delta_j^i < \Delta_{\mathcal{I}}\} \dfrac{26}{(\Delta_\mathcal{I})^2} \ln\Biggl(\dfrac{16\pi\sqrt{\dfrac{N}{3\delta}}}{(\Delta_{\mathcal{I}})^2}\Biggr)+1\Biggr).$$
\end{theorem}

We make the following remarks on the high probability upper bound on the stopping time under Algorithm~\ref{algo: vanilla}. First, the dependence on the vector of required arms~$r$ is implicit through the definition of the bottleneck distance~$\Delta_{\mathcal{I}}$. Secondly, the expression includes a summation over all the arms, each term capturing an upper bound on the number of rounds that arm is pulled. Intuitively, arms for which the gap~$\Delta_j^i$ (as defined in Definition~\ref{def: arm gap}) exceeds the bottleneck gap $\Delta_{\mathcal{I}}$ are most likely to be assigned to their respective clusters by the algorithm; the number of rounds these arms remain active is (with high probability) inversely proportional to $(\Delta_j^i)^2.$ On the other hand, arms whose gap~$\Delta_j^i$ is less than the bottleneck gap $\Delta_{\mathcal{I}}$ are most likely to remain active (i.e., not assigned to their respective clusters) until the algorithm stops; the number of rounds the algorithm needs to terminate being (with high probability) inversely proportional to $(\Delta_{\mathcal{I}})^2.$ Finally, note that this bound is `order-wise' consistent with the information theoretic lower bound in Theorem~\ref{theorem: lower bound}; both bounds exhibit an inverse proportionality to the square of the bottleneck gap. This suggests that the Vanilla Round Robin Algorithm is near optimal.\footnote{The lower bound in Theorem~\ref{theorem: lower bound} is on the \emph{expected} stopping time, whereas the upper bound in Theorem~\ref{theorem: vanilla} holds with \emph{high probability}; these quantities are not comparable strictly speaking. This `abuse' is however standard in the analysis of confidence interval based MAB algorithms (see~\cite{jamieson2014best}).}

\noindent {\bf Butterscotch Round Robin Algorithm for RAI}

Next, we discuss the \textit{Butterscotch Round Robin Algorithm}, stated formally as Algorithm \ref{algo: butterscotch}. The key difference from the Vanilla Algorithm discussed before is that the arm pulls are conducted in batches (line 5). Specifically, in round $R$, each arm in the active set $A$ is pulled $t_R - t_{R -1}$ times, where the value of $t_R$ is chosen (line 2) so that the confidence interval for the mean reward of each arm in $A$ is of size $\propto 2^{-R}.$  
This is reflected in the confidence-interval based cluster membership check conducted in line 10. The rest of Algorithm \ref{algo: butterscotch} proceeds in essentially the same way as Algorithm \ref{algo: vanilla}.  

\begin{algorithm}[t]
\caption{Butterscotch Round Robin Algorithm for RAI}
\label{algo: butterscotch}
\textbf{Input}: cluster sizes $c = (c_1, c_2, \cdots, c_m)$, required arms $r = (r_1, r_2, \cdots, r_m)$, arm set $\mathcal{N}$, error threshold~$\delta$ \\
\textbf{Output}: $O_1, O_2, \cdots, O_m$
\begin{algorithmic}[1]
\STATE Initialize $R \leftarrow 0, A \leftarrow \mathcal{N},$ and for $i \in \{1, 2, \cdots, m\}$ set $\tilde{c}_i = c_i$ 
\STATE Set $t_0 = 0$ and $t_R = 2^{(2R+5)}\ln\Bigl(\dfrac{\pi^2R^2N}{3\delta} \Bigr)$
\WHILE{$|O_1| \neq r_1 \text{ or } |O_2| \neq r_2 \text{ or } \cdots |O_m| \neq r_m$}
\STATE Increment $R$ by $1$
\STATE Sample every arm in $A$ $t_R - t_{R-1}$ times 
\STATE Update empirical mean rewards
\STATE 
Partition~$A$ into clusters $A_1, A_2, \cdots, A_m$ of sizes $\tilde{c}_1, \tilde{c}_2, \cdots, \tilde{c}_m$ respectively, based on the empirical means
\FOR{$i$ in $[m]$}
\FOR{arm $a$ in $A_i$}
\STATE $\mathrm{Better}(a) = \left\{j \in \cup_{j < i} A_j \colon \hat{\mu}_j - \hat{\mu}_a > 2^{-(R+2)} \right\}$ \\
\STATE $\mathrm{Worse}(a) = \left\{j \in \cup_{j > i} A_j \colon \hat{\mu}_a - \hat{\mu}_j > 2^{-(R+2)}  \right\}$ \\
\IF {$|\mathrm{Better}(a)| \geq \sum_{j < i} \tilde{c}_j$ and $|\mathrm{Worse}(a)| \geq \sum_{j > i} \tilde{c}_j$}
\IF{$|O_i| < r_i$}
\STATE Add arm~$a$ to $O_i$
\ENDIF
\STATE Remove arm~$a$ from $A$ 
\STATE $\tilde{c}_i \leftarrow \tilde{c}_i - 1$
\ENDIF
\ENDFOR
\IF{$i > 1$ and $|O_{i-1}| = r_{i-1}$ and $|O_{i}| = r_{i}$}
\STATE $\tilde{c}_i \leftarrow \tilde{c}_{i-1} + \tilde{c}_i$
\STATE $\tilde{c}_{i-1} \leftarrow 0$
\ENDIF
\ENDFOR
\ENDWHILE
\end{algorithmic}
\end{algorithm}
The structure of Algorithm \ref{algo: butterscotch} is inspired by the scheme proposed in \cite{karnin2013almost} for the best arm identification problem, where it is shown that the batch property enables tighter confidence intervals and consequently a better upper bound on the sample complexity as compared to standard successive elimination based schemes. We find that something similar is true for the RAI problem as well, as illustrated by the following result and the discussion thereafter.

\begin{theorem}
\label{theorem: butterscotch}
    The Butterscotch Round Robin Algorithm (see Algorithm~\ref{algo: butterscotch}) is $\delta$-PC for the RAI problem. With probability at least $1-\delta$, its sample complexity $\mathcal{T}_\delta^{\mathcal{I}}$ satisfies
    \begin{align*}
       \mathcal{T}_\delta^{\mathcal{I}} \le \sum_{i=1}^{m} \sum_{j = 1}^{c_i}\Biggl(\mathbb{1}\{\Delta_j^i \geq \Delta_{\mathcal{I}}\}\max\Biggl({ \dfrac{32}{(\Delta_j^i)^2} \ln \Biggl(\dfrac{N\pi^2}{3\delta} \ceil[\Bigg]{\log_2 \Biggl(\dfrac{1}{2\Delta_j^i}\Biggr)}^2\Biggr)},  128 \ln \Biggl( \dfrac{N\pi^2}{3\delta}\Biggr)\Biggl) \\ + \mathbb{1}\{\Delta_j^i < \Delta_{\mathcal{I}}\} \max\Biggl({\dfrac{32}{(\Delta_{\mathcal{I}})^2} \ln \Biggl(\dfrac{N\pi^2}{3\delta} \ceil[\Bigg]{\log_2 \Biggl(\dfrac{1}{2\Delta_{\mathcal{I}}} \Biggr)}^2 \Biggr) \Biggr)},  128 \ln \Biggl( \dfrac{N\pi^2}{3\delta}\Biggr)\Biggr).
    \end{align*}
\end{theorem}

The bound above has a similar form to the one in Theorem \ref{theorem: vanilla}. The main difference is that while the former has a $\log(1 / (\text{arm gap}))$ (or $\log ( 1 / \Delta_{\mathcal{I}}))$ dependence, the latter has a term proportional to only $\log\log(1 / (\text{arm gap}))$ (or $\log\log ( 1 / \Delta_{\mathcal{I}}))$. This potential improvement in sample complexity is due to the incorporation of batch pulling in Algorithm~\ref{algo: butterscotch}, which allows us to relax the union bound requirements in the proof. The empirical performance of the two algorithms is compared in Section~\ref{sec: numerical}.
The proofs for Theorems \ref{theorem: vanilla} and \ref{theorem: butterscotch} can be found in Appendix \ref{Theorem Proofs}. For each algorithm, the proof has two components: showing that it is $\delta$-PC and then deriving an upper bound on the sample complexity. 

\section{Numerical Case Studies}
\label{sec: numerical}
In this section, we conduct an empirical evaluation of Algorithms \ref{algo: vanilla} and \ref{algo: butterscotch} using both synthetic and real-world datasets. In addition, we also consider a suitably tailored version of LUCB-style sampling \cite{kalyanakrishnan2012pac}, which has been widely used in the multi-armed bandit literature and offers a sequential sampling strategy as opposed to the parallel nature of the successive elimination style strategies. In each round, the LUCB algorithm considers every empirical cluster for which the arms requirement hasn't yet been fulfilled and then based on the current confidence intervals of the mean estimates, selects from each cluster an arm (together with the `boundary' arms of the neighboring clusters) whose membership is most likely to be confirmed with the additional pull; further details along with a proof that the proposed variant is $\delta$-PC are provided in Appendix \ref{LUCB}. For our simulations, we assume the reward distribution to be Bernoulli$[0,1]$, a member of the $1/2$ SubGaussian distribution family. Finally, we set the error probability $\delta = .01$ and present sample complexity results which are averaged over $100$ independent runs of the corresponding algorithms. 

We first examine an instance with 10 arms divided into clusters of sizes $3, 5$, and $2$, respectively. The true means for this instance are given by: $[0.9, 0.85, 0.7, 0.66, 0.65, 0.6, 0.4, 0.35, 0.2, 0.15]$. Table \ref{tab: empirical comparisons} presents the average sample complexity of the three algorithms for various special cases of the RAI problem. 

\begin{table}[H]
\caption{Comparision of sample complexity between Algorithm \ref{algo: vanilla}, Algorithm \ref{algo: butterscotch}, and an LUCB-style scheme for various instantiations of the RAI problem}
    \centering
    \begin{tabular}{|c|c|c|c|c|c|}
    \hline 
     Sr.no & Task & Required Arms & Vanilla  & Butterscotch & LUCB \\   
     \hline \hline
     1 & One arm from top cluster & $(1,0,0)$ & 4588 & 10370 & 3634 \\ \hline
     2 & Identify top cluster & $(3,0,0)$ & 48632 & 31757 & 63802\\ \hline
     3 & $2$ arms each from top $2$ clusters & $(2,2,0)$ & 8958 & 10370 & 19444 \\ \hline
     4 & Identify center cluster & $(0,5,0)$ & 56730 & 50142 & 61302\\ \hline 
     5 & One arm from worst cluster & $(0,0,1)$ & 8913 & 10370 & 9257\\ 
     \hline 
    \end{tabular}
    \label{tab: empirical comparisons}
\end{table}

We observe that in almost all the cases, the Vanilla and Butterscotch schemes perform better than the LUCB-based algorithm. Intuitively, this is because the LUCB algorithm inherently targets sequential membership identification of arms which can lead to many unnecessary pulls for the boundary arms especially when the number of required arms is higher. Secondly, between the Vanilla and Butterscotch algorithms, the former performs better in Problems $1, 3, 5$ while the latter is superior for Problems $2, 4$. Problems $1, 3, 5$ turn out to be simpler problems that require only one round of the Butterscotch scheme (which is why the sample complexity is also the same for all three problems), while the Vanilla algorithm requires even fewer since it can stop at any time and does not have a batch constraint. 

Both Algorithms \ref{algo: vanilla} and \ref{algo: butterscotch} employ adaptive merging of clusters when the representative arm requirement for neighboring clusters is 
found to be already satisfied. To demonstrate the benefit of this feature, we conduct an empirical comparison of both algorithms with their corresponding non-merging versions, as shown in Table \ref{tab: merging comparisons}. The underlying bandit instance and the problems considered are the same as those in Table \ref{tab: empirical comparisons}.

\begin{table}[H]
\caption{Comparison of sample complexity between Algorithms \ref{algo: vanilla}, \ref{algo: butterscotch} and their versions which do not use adaptive merging of clusters}
    \label{tab: merging comparisons}
    \centering
    \begin{tabular}{|c|c|c|c|c|}
    \hline 
     Sr.no & Vanilla & Vanilla (Non Merging)  & Butterscotch & Butterscotch (Non Merging) \\   
     \hline \hline
     1 & 4588 & 6491 & 10370 & 10370 \\ \hline
     2 & 48632 & 50350 & 31757 & 44358\\ \hline
     3 & 8958 & 10745 & 10370 & 10370 \\ \hline
     4 & 55932 & 56125 & 50162 &  51002\\ \hline 
     5 & 8913 & 10393 & 10370 & 10370 \\ 
     \hline 
    \end{tabular}
\end{table}

\begin{figure}
    \centering
    \subfigure[]{\includegraphics[scale = 0.35]{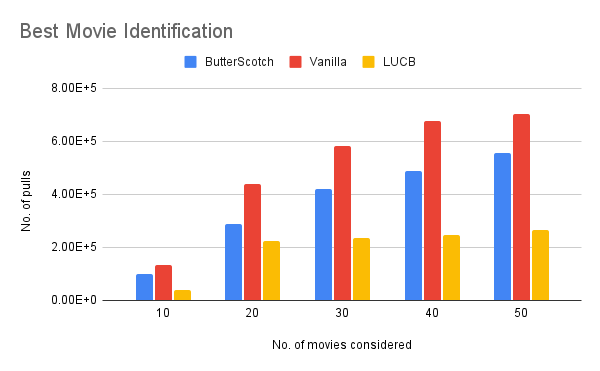}} 
    \subfigure[]{\includegraphics[scale = 0.35]{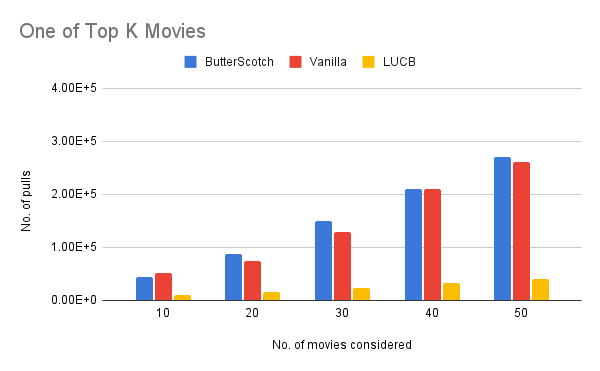}} 
    \subfigure[]{\includegraphics[scale = 0.35]{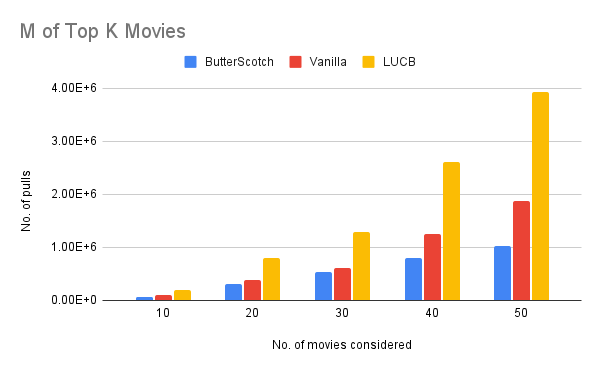}}
    \subfigure[]{\includegraphics[scale = 0.35]{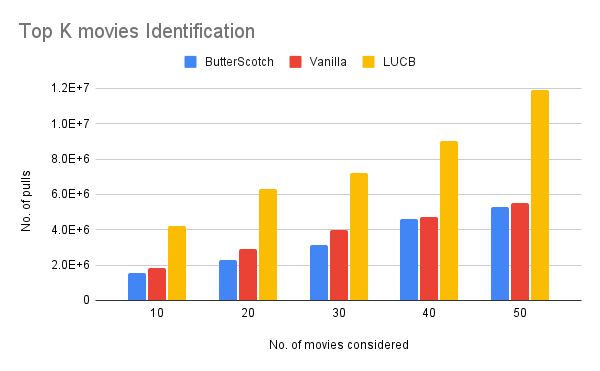}}
    \subfigure[]{\includegraphics[scale = 0.35]{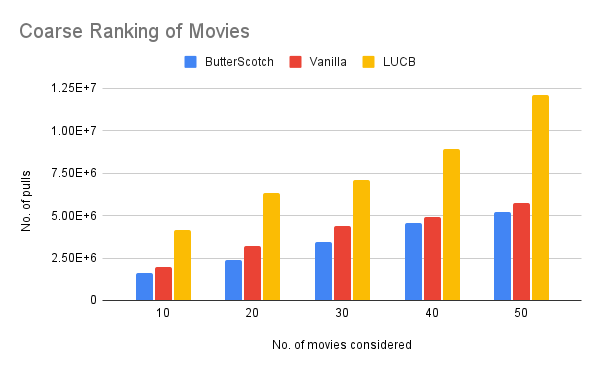}}
    \subfigure[]{\includegraphics[scale = 0.35]{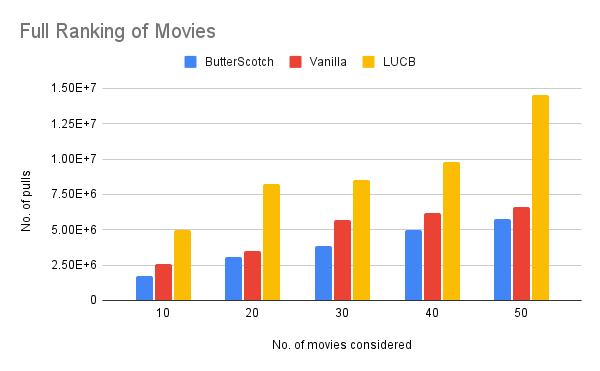}}
    \caption{Comparision of sample complexity between Algorithm \ref{algo: vanilla}, Algorithm \ref{algo: butterscotch}, and an LUCB-style scheme for special cases of the RAI problem, over an instance created from the MovieLens dataset}
    \label{fig: movielens}
\end{figure}

In addition to the synthetic dataset, we run our algorithms on the MovieLens dataset \cite{cantador2011second} which is a large database of movies and corresponding user ratings. Out of the approximately $27,000$ movies in the dataset, we shortlist a set of $50$ movies that have received the highest number of ratings. We consider the movies as arms and their average user rating (normalized to [0,1]) as the corresponding mean reward. Each time we pull the arm associated with a movie, a random reward is generated which corresponds to the rating from a randomly chosen user. We experiment with different values of arms ($N$) ranging from $10$ to $50$, and consider all the special cases of the RAI problem stated in Table \ref{tab: standard examples}. These include best arm identification; identifying $1$, $M$, and all of the top $K$ arms; and finally coarse and full ranking. We select $M$ and $K$ to be $20\%$ and $50\%$ of $N$, respectively, and for the coarse ranking problem, the clusters are divided in the ratio of $3:5:2$. The results are presented in Figure \ref{fig: movielens}. 

From Figures \ref{fig: movielens}(a),(b), the LUCB algorithm performs the best for the best arm and $1$ out of the top $K$ identification problems. This is in line with our earlier observation over the synthetic dataset in Table~\ref{tab: empirical comparisons} where we found LUCB to be superior when the number of required arms is smaller. For the other four problems depicted in Figures~\ref{fig: movielens}(c),(d),(e),(f), the Vanilla and Butterscotch algorithms are significantly better than LUCB. The Butterscotch algorithm almost always performs the best, while providing a sizeable advantage over the Vanilla algorithm in many cases. The hardness of the various problems under consideration is also easily visible from the results; for example, for the hardest problem, i.e., getting a full ranking of movies, the number of pulls required is the maximum and is much larger than the problem of identifying $1$ out of the top $K$ movies.

\section{Conclusion}
\label{sec: Conclusion}
We have proposed the representative arm identification (RAI) problem which generalizes several well studied problems in multi-armed bandits, including best arm identification, identifying $M$ out of the top $K$ arms, and coarse ranking. We provided a lower bound on the sample complexity of any reliable scheme and also proposed two algorithms based on the idea of confidence intervals. Upper bounds on the sample complexity of these algorithms were presented and their empirical performance was demonstrated over synthetic and real-world datasets. 

Some questions remain open and several directions can be pursued in the future:
\begin{enumerate}
\item While Theorem~\ref{theorem: lower bound} presents an instance-dependent and interpretable lower bound on the sample complexity of the RAI problem, it is in general loose. On the other hand, \cite{degenne2019pure} provides a  lower bound on problems with multiple correct answers (which includes the RAI problem) which can in general be tighter, but the bound is in the form of an optimization problem and is hard to compare against. Deriving a tighter lower bound which is still interpretable in terms of the problem parameters is a key open problem and can provide insights into multiple problems of interest given the many special cases that RAI covers.

\item \textit{Federated RAI}: In this version of the problem, we have multiple clients, each having its own mean reward vector corresponding to the arms. Each client aims to solve its own local RAI problem, while a server which can communicate with the clients (at some cost) might be interested in solving a global RAI problem. A preliminary study of the problem indicates that a carefully crafted combination of the two algorithms proposed in this work can be used to solve the federated RAI problem, while ensuring low communication cost. 

\end{enumerate}

\bibliographystyle{unsrtnat}  
\bibliography{main}  


\newpage
\appendix

\section{Proof of Theorems}
\label{Theorem Proofs}

Before proceeding with the proofs, let us state the necessary concentration inequality used in proving Theorem \ref{theorem: vanilla} and Theorem \ref{theorem: butterscotch}

\begin{lemma}
    \label{lemma: inequality}
    Let $X_1, X_2, \cdots, X_R \in \frac{1}{2}-$SubGaussian be independent and identical random variables. Than for the empirical mean $\hat{\mu}(R) = \sum_{i=1}^R X_i/R$ we have
    \begin{align*}
    P(|\hat{\mu}(R) - \mu| \geq \epsilon_R) \leq 2 \exp{(-2R\epsilon_R^2)}
    \end{align*}
\end{lemma}
\textbf{Proof of Lemma \ref{lemma: inequality}.} 
From Hoeffdings inequality for any $\sigma^2-$SubGaussian random variable, we have
\begin{align}
\label{hoeffding}
P(|\hat{\mu}(R) - \mu| \geq \epsilon_R) \leq 2 \exp{\left( - \dfrac{R \epsilon_R^2}{2 \sigma^2}\right)}
\end{align}
On substituting $\sigma = \frac{1}{2}$ in Equation \ref{hoeffding}, for any $\frac{1}{2}-$SubGaussian random variable, we get
\begin{align}
P(|\hat{\mu}(R) - \mu| \geq \epsilon_R) \leq 2 \exp{\left( - 2 R \epsilon_R^2\right)} \label{subgaussian}
\end{align}
\subsection{Proof of Theorem \ref{theorem: lower bound}}
\label{Proof of lower bound}
The proof of the lower bound on expected sample complexity starts from recognizing that the RAI problem fits in the `multiple correct answers' framework of \cite{degenne2019pure} and thus the general lower bound derived there applies to the RAI problem as well. However, that lower bound is in the form of a $\min \min \max$ optimization problem and the rest of the proof involves simplifying it to obtain an interpretable form. 

To state the lower bound in \cite{degenne2019pure}, we have to introduce some additional notation. Consider an RAI problem instance $\mathcal{I} = (c, r, \Pi)$, where the arm reward distributions are Gaussian with standard deviation $1/2$ and the mean reward vector is given by $\mu := \{\mu_i^j\}$, where $\mu_i^j$ is the mean reward for arm $j$ from cluster $i$ under the current instance. Let $i^*[\mu]$ denote the set of all correct answers when the reward distributions are specified by $\mu$. Note that each such correct answer corresponds to a set of arms such that $r_1$ of them belong to cluster $1$, $r_2$ of them belong to cluster $2$ and so on. Next, for any correct answer $a \in i^*[\mu]$, we need the notion of an \textit{alternate mean reward vector} $\lambda$ such that $a$ is not a correct answer when the underlying arm reward distributions are Gaussian with standard deviation $1/2$ and the mean reward vector is $\lambda = \{\lambda_i^j\}$. We will denote the collection of all such alternate mean reward vectors by $\neg a$. Finally, let $\Delta_K$ denote the $K$-dimensional simplex and $d(a,b)$
denote the Kullback-Leibler (KL) divergence between two Gaussian distributions with means $a$ and $b$, and variance $1/2$ each. Note that $d(a,b) = 2(a - b)^2$.

Next, from \citep[Theorem 1]{degenne2019pure}, we have the following lower bound on the expected sample complexity of any $\delta$-PC algorithm for an RAI problem $\mathcal{I} = (c, r, \Pi)$, where the arm reward distributions are Gaussian with variance $1/2$ and the mean reward vector is given by $\mu := \{\mu_i^j\}$:
\begin{align}
\label{mca}
\liminf_{\delta \ra 0} \dfrac{E[T^{\mathcal{I}}_{\delta}(\mathcal{A})]}{\log(1/\delta)} \geq D(\mathcal{I})^{-1}
\end{align}
where $D(\mathcal{I}) = \max\limits_{\substack{a \in i^*[\mu]}} \max\limits_{\substack{w \in \Delta_N}} \inf\limits_{\substack{\lambda \in \neg a}} \sum_{i=1}^m \sum_{j=1}^{c_i} w_j^i d(\mu_j^i, \lambda_j^i).$

\begin{figure}[H]
    \centering
    \includegraphics[scale = 0.5]{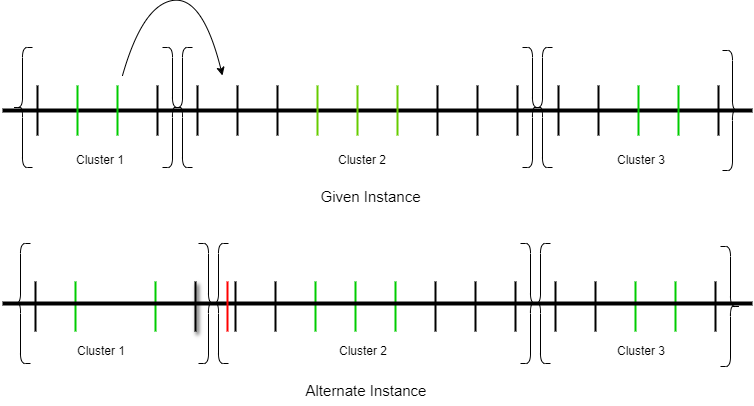}
    \caption{An illustration of alternate instance. In the given instance, the arms marked in green represent a set of correct answers. Now, to form an alternate instance, we shift an arm from cluster 1 to cluster 2, placing it just behind the boundary of cluster 2 in the given instance.}
    \label{fig: lowerbound}
\end{figure}

Now, we will derive an upper bound on  $D(\mathcal{I})$, which will yield the lower bound in the expression of Theorem~\ref{theorem: lower bound}. Consider any given correct answer $a \in i^*[\mu]$. Then, a particular alternate mean reward vector $\lambda \in \neg a$, can be constructed by shifting the mean reward for any one arm included in $a$, say arm $k$, so that its cluster membership is changed, and keeping all other mean rewards the same; see Figure \ref{fig: lowerbound} for an illustration. In particular, the minimum shift needed to do so is given by the arm gap for arm $k$, as defined in Definition~\ref{def: arm gap}. Amongst all the arms included in $a$, we will choose the one, say $b_a$, which requires the smallest change in mean reward to result in an alternate mean reward vector, i.e., for which the set of arms $a$ is no longer a correct answer. Denote the corresponding change in the mean reward of arm $b_a$ by $l_a$. Then we have 
\begin{align*}
D(\mathcal{I})  &=  \max\limits_{\substack{a \in i^*[\mu]}}  \max\limits_{\substack{w \in \Delta_N}} \inf\limits_{\substack{\lambda \in \neg a}} \sum_{i=1}^m \sum_{j=1}^{c_i} w_j^i d(\mu_j^i, \lambda_j^i) \\ &\leq \max\limits_{\substack{a \in i^*[\mu]}} \max\limits_{\substack{w \in \Delta_N}} w_{b_a} 2l_a^2 \\
   &= \max\limits_{\substack{a \in i^*[\mu]}} 2l_a^2 \\
   &= 2 (\Delta_{\mathcal{I}})^2
\end{align*}
where the last equality follows by recognizing that amongst all the correct answers $a \in i^*[\mu]$, the one that will need the biggest change in the mean reward of an arm to create an alternate reward distribution will be the one which contains the following: $r_1$ arms from cluster $1$ with the highest arm gaps, $r_2$ arms from cluster $2$ with the highest arm gaps, and so on; and the corresponding change will be equal to $\Delta_{\mathcal{I}}$ by Definition~\ref{def: optimal gap}. Substituting the above bound in \eqref{mca} completes the proof of Theorem~\ref{theorem: lower bound}.
\subsection{Proof of Theorem \ref{theorem: vanilla}} 
\label{Proof of Vanilla}
The proof of this theorem is organized as follows. First, we define a good event $\psi$. Based on this condition, we then demonstrate the correctness of Claims \ref{claim: vanilla correctness} and \ref{claim: vanilla complexity}.
\begin{align}
\label{good event vanilla}
    \psi \triangleq \left\{\forall R \in Z, \forall i \in [m], \forall j \in [c_i], |\hat{\mu}_{j}^i(R) - \mu_{j}^i| \leq \sqrt{\frac{\ln(\pi^2R^2N/3\delta)}{2R}} \right\}
\end{align}
\begin{claim}
    \label{claim: vanilla correctness}
    Under $\mathcal{A} = Alg1$, let $\hat{\mu}_{j}^i(R)$ be the empirical mean of arm $j$ from cluster $i$ at round $R$. Let $\psi$ be the event defined in \ref{good event vanilla}. Than the $P_{\mathcal{I}}^{\mathcal{A}}(\psi) \geq 1-\delta$
\end{claim}
\textbf{Proof of Claim \ref{claim: vanilla correctness}.} By the union bound, we have
\begin{align}
\label{vanilla correctness}
    P_{\mathcal{I}}^{\mathcal{A}}(\psi^{c}) &\leq \sum_{R=1}^\infty \sum_{i=1}^{m} \sum_{j = 1}^{c_i} P_{\mathcal{I}}^{\mathcal{A}}\left[|\hat{\mu}_{j}^i(R) - \mu_{j}^i| > \sqrt{\frac{\ln(\pi^2R^2N/3\delta)}{2R}}\right] 
\end{align}
    From Equation \ref{subgaussian} and \ref{vanilla correctness}, we have
\begin{align}
    P_{\mathcal{I}}^{\mathcal{A}}(\psi^{c}) &\leq \sum_{R=1}^\infty \sum_{i=1}^{m} \sum_{j = 1}^{c_i} \dfrac{6\delta}{\pi^2R^2N}   \\
    &\leq \sum_{R=1}^\infty\dfrac{6\delta}{\pi^2R^2} \\
    &\leq \delta
\end{align}
\begin{lemma}
    \label{lemma: vanilla complexity}
    Let $\mathcal{T}_{i,j}$ be the time step of $Alg1$ in which the identity of the arm $j$ from cluster $i$ is identified, conditioned on the good event $(\psi)$, we have $\mathcal{T}_{i,j} \leq \mathcal{T}_{i,j}^{\mathcal{I}}$ where $\mathcal{T}_{i,j}^{\mathcal{I}}$ is defined as
    \begin{align*}
        \mathcal{T}_{i,j}^{\mathcal{I}} &= \dfrac{26}{{\Delta_j^i}^2} \ln\Biggl(\dfrac{16\pi\sqrt{\dfrac{N}{3\delta}}}{ {\Delta_j^i}^2}\Biggr)+1
    \end{align*}
\end{lemma}



\textbf{Proof of Lemma \ref{lemma: vanilla complexity}.} 
The proof of Lemma \ref{lemma: vanilla complexity} follows a similar approach to that of Lemma 3 in \cite{reddy2022almost}, differing only in the confidence intervals. However, for the reader's convenience, we provide the complete proof of the lemma here.

Consider the $Alg1$ identifies an arm $j$ from cluster $i$ in time step $\mathcal{T}_{i,j}$, then it must be the case that on the good event $\psi$, the following holds true
\begin{align*}
    \min\left\{\hat{\mu}_{\text{arm}}^{i}(\mathcal{T}_{i,j}) - \hat{\mu}_{1}^{i+1}(\mathcal{T}_{i,j}), \hat{\mu}_{\tilde{c}_{i-1}}^{i-1}(\mathcal{T}_{i,j}) - \hat{\mu}_{\text{arm}}^{i}(\mathcal{T}_{i,j})\right\} \geq 2\sqrt{\frac{\ln(\pi^2\mathcal{T}_{i,j}^2N/3\delta)}{2\mathcal{T}_{i,j}}}
\end{align*}
Note that $\sqrt{\dfrac{\ln(\pi^2R^2N/3\delta)}{2R}} \rightarrow 0$ as $R \rightarrow \infty$.

Now, let $R_j^i \coloneqq \inf \left\{R: \sqrt{\dfrac{\ln(\pi^2R'^2N/3\delta)}{2R'}} \leq \Delta^i_j/4 \; \forall R' \geq R\right\}$, then it must be that 
\begin{align}
\label{intermediate lemma 2}
    \min\left\{\hat{\mu}_{\text{arm}}^i(R) - \hat{\mu}_{1}^{i+1}(R), \hat{\mu}_{\tilde{c}_{i-1}}^{i-1}(R) - \hat{\mu}_\text{{arm}}^{i}(R)\right\} \geq 2\sqrt{\frac{\ln(\pi^2R^2N/3\delta)}{2R}}, \; \forall R > R_j^i
\end{align}
From equation \ref{intermediate lemma 2}, it follows that on the good event $\psi$, $\mathcal{T}_{i,j} < R_j^i$. We now derive an upper bound on $R_j^i$ by letting
\begin{align*}
    {R_j^i}' \coloneqq \left\lceil \max \left\{ n \in (1, \infty): \sqrt{\frac{\ln(\pi^2n^2N/3\delta)}{2n}} = \Delta^i_j/4 \right\}\right\rceil
\end{align*}
The maximum in the above equation picks the largest solution for $n\in(1, \infty)$ satisfying $\sqrt{\frac{\ln(\pi^2n^2N/3\delta)}{2n}} = \Delta^i_j/4$, while ceil returning the smallest integer corresponding to $n$. We clearly see that $R_j^i \leq {R_j^i}'$. 

Now, the exact expression for ${R_j^i}'$ is given by ${R_j^i}' = \left\lceil -\frac{1}{a} W_{-1}(-ae^{-b})\right\rceil$ where for $y < 0$, $W_{-1}(y)$ is the smallest value of $x$ such that $xe^x = y$. Note that here $W_{-1}(y)$ refers to the Lambert W function \cite{reddy2022almost}

Note that while all these steps are followed from \cite{reddy2022almost}, due to the difference in the confidence intervals in our case 
\begin{align}
\label{ab}
    a = \dfrac{{\Delta^i_j}^2}{16} \text{ and } b = \dfrac{\ln\left( \dfrac{\pi^2N}{3\delta} \right)}{2}
\end{align}
From \cite{alzahrani2018sharp}, we also have 
\begin{align}
\label{lambert}
    W_{-1}(y) > \dfrac{e}{e-1} \ln(-y)
\end{align}
Therefore, from equation \ref{ab} and \ref{lambert}, we have
\begin{align*}
    \mathcal{T}_{i,j} &\leq {R_j^i}' \\
    &\leq \left\lceil \dfrac{e}{e-1} \dfrac{b-\ln(a)}{a}\right\rceil \\
    &\leq \dfrac{e}{e-1} \dfrac{b-\ln(a)}{a} + 1 \\
    &= \dfrac{16e}{e-1} \dfrac{1}{{\Delta^i_j}^2}\ln\Biggl(\dfrac{16\pi\sqrt{\dfrac{N}{3\delta}}}{ {\Delta_j^i}^2}\Biggr)+1 \\
    &= \dfrac{26}{{\Delta_j^i}^2} \ln\Biggl(\dfrac{16\pi\sqrt{\dfrac{N}{3\delta}}}{ {\Delta_j^i}^2}\Biggr)+1
\end{align*}

\begin{claim}
\label{claim: vanilla complexity}
Conditioned on the good event $\psi$, the $Alg1$ takes no more than $\mathcal{T}_\delta^{\mathcal{I}}(Alg1)$ pulls to the solve the representative identification problem where
\begin{align*}
\mathcal{T}_\delta^{\mathcal{I}}(Alg1) \le  \sum_{i=1}^{m} \sum_{j = 1}^{c_i}\Biggl(\mathbb{1}\{\Delta_j^i \geq \Delta_{\mathcal{I}}\} \dfrac{26}{{\Delta_j^i}^2} \ln\Biggl(\dfrac{16\pi\sqrt{\dfrac{N}{3\delta}}}{ {\Delta_j^i}^2}\Biggr) + \mathbb{1}\{\Delta_j^i < \Delta_{\mathcal{I}}\} \dfrac{26}{\Delta_{\mathcal{I}}^2} \ln\Biggl(\dfrac{16\pi\sqrt{\dfrac{N}{3\delta}}}{\Delta_{\mathcal{I}}^2}\Biggr)+1\Biggr)
\end{align*}
\end{claim}
\textbf{Proof of Claim \ref{claim: vanilla complexity}.} Given the number of arms $(N)$, the error probability $(\delta)$, and the arm gap $(\Delta_j^i)$, from Lemma \ref{lemma: vanilla complexity}, we know that for any arm $j$ from cluster $i$, conditioned on the good event $\psi$, the maximum number of pulls for the arm by $Alg1$ is no more than $\mathcal{T}_{i,j}^{\mathcal{I}}$, where
\begin{align}
    \mathcal{T}_{i,j}^{\mathcal{I}} = \dfrac{26}{{\Delta_j^i}^2} \ln\Biggl(\dfrac{16\pi\sqrt{\dfrac{N}{3\delta}}}{ {\Delta_j^i}^2}\Biggr)+1
\end{align}
However, as at every round $R$, the algorithm pulls all the unidentified arms, the total number of pulls by $Alg1$ will be equivalent to the sum of the individual pulls.
Now, from the definition of the Bottleneck gap (Refer Definition \ref{def: optimal gap}), we know that by the time the arm corresponding to the bottleneck gap gets identified, the cluster requirement would be satisfied resulting in the stopping of the algorithm. This leaves us with the $2$ cases.

Case 1: When $\Delta_j^i \geq \Delta_\mathcal{I}$\\
In this case, the number of pulls $T_{i,j}^\mathcal{I}$ for arm $j$ from cluster $i$ is 
\begin{align}
\label{case1}
    \mathcal{T}_{i,j}^{\mathcal{I}} = \dfrac{26}{{\Delta_j^i}^2} \ln\Biggl(\dfrac{16\pi\sqrt{\dfrac{N}{3\delta}}}{ {\Delta_j^i}^2}\Biggr)+1
\end{align}

Case 2: When $\Delta_j^i < \Delta_\mathcal{I}$
For this case, as the algorithm terminates before identifying the identity of arm $j$ from cluster $i$, we have
\begin{align}
\label{case3}
    \mathcal{T}_{i,j}^{\mathcal{I}} = \dfrac{26}{\Delta_\mathcal{I}^2} \ln\Biggl(\dfrac{16\pi\sqrt{\dfrac{N}{3\delta}}}{ {\Delta_\mathcal{I}}^2}\Biggr)+1
\end{align}
Finally, on summing up the number of pulls for every arm, we have
\begin{align*}
\mathcal{T}_\delta^{\mathcal{I}}(Alg1) \le  \sum_{i=1}^{m} \sum_{j = 1}^{c_i}\Biggl(\mathbb{1}\{\Delta_j^i \geq \Delta_{\mathcal{I}}\} \dfrac{26}{{\Delta_j^i}^2} \ln\Biggl(\dfrac{16\pi\sqrt{\dfrac{N}{3\delta}}}{ {\Delta_j^i}^2}\Biggr) + \mathbb{1}\{\Delta_j^i < \Delta_{\mathcal{I}}\} \dfrac{26}{\Delta_{\mathcal{I}}^2} \ln\Biggl(\dfrac{16\pi\sqrt{\dfrac{N}{3\delta}}}{\Delta_{\mathcal{I}}^2}\Biggr)+1\Biggr)
\end{align*}

\subsection{Proof of Theorem \ref{theorem: butterscotch}:} 
\label{Proof of Butterscotch}
The proof of Theorem \ref{theorem: butterscotch}, follows a similar approach to that of Appendix \ref{Proof of Vanilla}. We start with defining the good event, conditioned on which the rest of the proof follows.
\begin{align}
\label{good event butterscotch}
    \psi \triangleq \left\{\forall R \in Z, \forall i \in [m], \forall j \in [c_i], |\hat{\mu}_{j}^i(R) - \mu_{j}^i| \leq 2^{-(R+3)} \right\}  
\end{align}
\begin{claim}
\label{claim: butterscotch correctness}
    Under $\mathcal{A} = Alg2$, let $\mu_j^i$ be the empirical mean of arm $j$ from cluster $i$ after $t_R$ pulls, where $t_R$ is defined in the line 2 of the $Alg2$. Given $\psi$ be the good event defined in \ref{good event butterscotch}, we have $P_{\mathcal{I}}^{\mathcal{A}}(\psi) \geq 1-\delta$
\end{claim}
\textbf{Proof of Claim \ref{claim: butterscotch correctness}.} From union bounding, we have
\begin{align}
\label{butterscotch correctness}
    P_{\mathcal{I}}^{\mathcal{A}}(\psi^{c}) &\leq \sum_{R=1}^\infty \sum_{i=1}^{m} \sum_{j = 1}^{c_i} P_{\mathcal{I}}^{\mathcal{A}}\left[|\hat{\mu}_{j}^i(R) - \mu_{j}^i| > \frac{2^{-R}}{8}\right] 
\end{align}
Now, on using the equations \ref{subgaussian} and \ref{butterscotch correctness}, and from line 2 of the Algorithm \ref{algo: butterscotch}, we have
\begin{align}
    P_{\mathcal{I}}^{\mathcal{A}}(\psi^{c}) &\leq \sum_{R=1}^\infty \sum_{i=1}^{m} \sum_{j = 1}^{c_i} \dfrac{6\delta}{\pi^2R^2N}   \\
    &\leq \sum_{R=1}^\infty\dfrac{6\delta}{\pi^2R^2} \\
    &\leq \delta
\end{align}
\begin{lemma}
\label{lemma: butterscotch complexity}
    Given the good event, to identify the identity of arm $j$ from cluster $i$, the $Alg2$, takes no more than $\mathcal{T}_{i,j}^{\mathcal{I}}$ pulls, where $\mathcal{T}_{i,j}^{\mathcal{I}}$ is defined as
    \begin{align*}
        \mathcal{T}_{i,j}^{\mathcal{I}} = \max\Biggl({ \dfrac{32}{{\Delta_j^i}^2} \ln \Biggl(\dfrac{N\pi^2}{3\delta} \ceil[\Bigg]{\log_2 \Biggl(\dfrac{1}{2\Delta_j^i}\Biggr)}^2\Biggr)},  128 \ln \Biggl( \dfrac{N\pi^2}{3\delta}\Biggr)\Biggl)
    \end{align*}
\end{lemma}
\textbf{Proof of Lemma \ref{lemma: butterscotch complexity}.} 
Following the argument in Lemma \ref{lemma: vanilla complexity}, we know that an arm $j$ from cluster $i$ will be identified with certainty at round $R$ if $2^{-(R+3)} = \Delta_j^i / 4$. Simplifying this further, we obtain:
\begin{align}
\label{R}
    R = \left \lceil \log_2 \left( \dfrac{1}{2\Delta_j^i} \right) \right \rceil
\end{align}
Next, from the definition of $t_R$ and the value of R from equation\ref{R}, we have
\begin{align*}
    \mathcal{T}_{i,j}^{\mathcal{I}} &= \ln\left(\dfrac{\pi^2N}{3\delta} \left \lceil \log_2 \left( \dfrac{1}{2\Delta_j^i} \right) \right \rceil^2\right) 2^{\left( 2 \left \lceil \log_2 \left( \dfrac{1}{2\Delta_j^i} \right) \right \rceil + 5\right)}\\
    &\leq \ln\left(\dfrac{\pi^2N}{3\delta} \left \lceil \log_2 \left( \dfrac{1}{2\Delta_j^i} \right) \right \rceil^2\right) 2^{\left( 2 \left(\log_2 \left( \dfrac{1}{2\Delta_j^i} \right) + 1 \right) + 5\right)} \\
    &= \dfrac{128}{4{\Delta_j^i}^2} \ln\left(\dfrac{\pi^2N}{3\delta} \left \lceil \log_2 \left( \dfrac{1}{2\Delta_j^i} \right) \right \rceil^2\right) \\
    &= \dfrac{32}{{\Delta_j^i}^2} \ln\left(\dfrac{\pi^2N}{3\delta} \left \lceil \log_2 \left( \dfrac{1}{2\Delta_j^i} \right) \right \rceil^2\right)
\end{align*}
However, since there are no pulls in round $0$, it's essential to consider the total number of pulls at round $1$ as part of the worst-case bound. Therefore, for any arm $j$, from cluster $i$, the total number of pulls will be bounded by
\begin{align*}
    \mathcal{T}_{i,j}^{\mathcal{I}} = \max\Biggl({ \dfrac{32}{{\Delta_j^i}^2} \ln \Biggl(\dfrac{N\pi^2}{3\delta} \ceil[\Bigg]{\log_2 \Biggl(\dfrac{1}{2\Delta_j^i}\Biggr)}^2\Biggr)},  128 \ln \Biggl( \dfrac{N\pi^2}{3\delta}\Biggr)\Biggl)
\end{align*}
\begin{claim}
\label{claim: butterscotch complexity}
Conditioned on the good event $\psi$, the $Alg2$ takes no more than $\mathcal{T}_\delta^{\mathcal{I}}(Alg2)$ pulls to the solve the representative identification problem where
\begin{align*}
    \mathcal{T}_\delta^{\mathcal{I}}(Alg2) \le \sum_{i=1}^{m} \sum_{j = 1}^{c_i}\Biggl(\mathbb{1}\{\Delta_j^i \geq \Delta_{\mathcal{I}}\}\max\Biggl({ \dfrac{32}{{\Delta_j^i}^2} \ln \Biggl(\dfrac{N\pi^2}{3\delta} \ceil[\Bigg]{\log_2 \Biggl(\dfrac{1}{2\Delta_j^i}\Biggr)}^2\Biggr)},  128 \ln \Biggl( \dfrac{N\pi^2}{3\delta}\Biggr)\Biggl) \\ + \mathbb{1}\{\Delta_j^i < \Delta_{\mathcal{I}}\} \max\Biggl({\dfrac{32}{\Delta_{\mathcal{I}}^2} \ln \Biggl(\dfrac{N\pi^2}{3\delta} \ceil[\Bigg]{\log_2 \Biggl(\dfrac{1}{2\Delta_{\mathcal{I}}} \Biggr)}^2 \Biggr) \Biggr)},  128 \ln \Biggl( \dfrac{N\pi^2}{3\delta}\Biggr)\Biggr)
\end{align*}
\end{claim}
\textbf{Proof of Claim \ref{claim: butterscotch complexity}.} The proof of Claim \ref{claim: butterscotch complexity} follows a similar argument as that of Claim \ref{claim: vanilla complexity}

\section{LUCB}
\label{LUCB}

The LUCB version for the representative identification problem is inspired from \cite{kalyanakrishnan2012pac} and is stated in Algorithm \ref{algo: lucb}. In this algorithm, we have an intelligent sampling rule following a pull strategy based on the LCB and UCB of the arms. We start this section, by defining the upper and lower confidence bounds.
\begin{definition} 
    \label{def: LCB UCB}
    (LCB and UCB of the arms): Given $\hat{\mu}_j^i(R)$ be the empirical mean of arm $j$ from cluster $i$, we define $\text{LCB}_j^i$ at round $R$ as 
    $$\text{LCB}_j^i (R) = \hat{\mu}_j^i (R) - \sqrt{\dfrac{\ln (\pi^2R^3N/3\delta)}{2R_j^i}}$$
    and $\text{UCB}_j^i$at round $R$ as 
    $$\text{UCB}_j^i (R) = \hat{\mu}_j^i (R)+ \sqrt{\dfrac{\ln (\pi^2R^3N/3\delta)}{2R_j^i}}$$
    where $R_j^i$ is the number of times the arm $j$ from cluster $i$ till round $R$
\end{definition}

Next, we identify the empirical gaps corresponding to every arm. These empirical gaps are defined in terms of the confidence bounds and play an important role in identifying the potential arms to pull.

\begin{definition} 
    \label{def: Empirical Gaps}
    (Empirical Gap): For all clusters $i \in [1, 2, \cdots, m]$, let $\text{LCB}^i(R)$  and $\text{UCB}^i(R)$ be
    $$\text{LCB}^i(R) = \min(\text{LCB}_1^i(R), \cdots \text{LCB}_{c_i}^i(R))$$ and $$\text{UCB}^i(R) = \max(\text{UCB}_1^i(R), \cdots \text{UCB}_{c_i}^i(R))$$
    than the empirical gap $\hat{\Delta}_j^i$ for arm $j$ from cluster $i$ is defined as
    $$\hat{\Delta}_j^i = \max(\text{UCB}_j^i(R)-\text{LCB}^{i-1}(R), \text{UCB}^{i+1}(R) - \text{LCB}_j^i(R)$$
\end{definition}

\begin{algorithm}
\caption{LUCB styled Algorithm for RAI}
\label{algo: lucb}
\textbf{Input}: cluster sizes $c = (c_1, c_2, \cdots, c_m)$, required arms $r = (r_1, r_2, \cdots, r_m)$, arm set $\mathcal{N}$, error threshold~$\delta$ \\
\textbf{Output}: $O_1, O_2, \cdots, O_m$
\begin{algorithmic}[1]
\STATE Initialize $R \leftarrow 0, A \leftarrow \mathcal{N},$ and for $i \in \{1, 2, \cdots, m\}$ set $\tilde{c}_i = c_i$ 
\STATE Sample every arm in $A$ once
\WHILE{$|O_1| \neq r_1 \text{ or } |O_2| \neq r_2 \text{ or } \cdots |O_m| \neq r_m$}
\STATE Increase $R$ by $1$
\STATE 
Partition~$A$ into clusters $A_1, A_2, \cdots, A_m$ of sizes $\tilde{c}_1, \tilde{c}_2, \cdots, \tilde{c}_m$ respectively, based on the empirical means
\FOR{$i$ in $[m]$}
\FOR{arm in $A_i$}
\STATE Calculate $\text{LCB}_{\text{arm}}^i (R)$ and $\text{UCB}_{\text{arm}}^i (R)$ 
\IF{$\text{LCB}(R)_\text{{arm}}^i > \text{UCB}(R)_{1}^{i+1}$ and $\text{UCB}_{\tilde{c}_{i-1}}^{i-1}(R) > \text{LCB}_\text{{arm}}^i(R)$} 
\IF{$|O_i| < r_i$}
\STATE Add arm to $O_i$
\ENDIF
\STATE Remove arm from $A$ 
\STATE $\tilde{c}_i \leftarrow \tilde{c}_i - 1$
\ENDIF
\ENDFOR
\IF{$O_i \neq r_i$}
\STATE $\text{LCB}^{i-1}(R) \leftarrow \min(\text{LCB}_1^{i-1}(R), \cdots \text{LCB}_{c_i}^{i-1}(R))$
\STATE $\text{UCB}^{i-1}(R) \leftarrow \max(\text{UCB}_1^{i-1}(R), \cdots \text{UCB}_{c_i}^{i-1}(R))$
\STATE $\hat{\Delta}^i(R) \leftarrow \min(\hat{\Delta}_1^i, \hat{\Delta}_2^i, \cdots \hat{\Delta}_{c_i}^i)$ 
\STATE Pull the arms corresponding to $\text{LCB}^{i-1}(R), \text{UCB}^{i-1}(R)$ and $\hat{\Delta^i}(R)$
\ENDIF
\ENDFOR
\ENDWHILE
\end{algorithmic}
\end{algorithm}

The algorithm starts by sampling every arm once (Line 2) before using a smarter pull-based strategy. It then enters a loop which continues until enough arms are identified from every cluster. While in the loop the algorithm first divides all the arms into clusters based on their empirical (line 5). It then calculates the LCB and the UCB corresponding to every arm, based on which the membership criteria is checked in Line 9. Finally, depending on whether the required number of arms are identified or not, it than decides if there are any additional pulls required for that cluster.

\begin{theorem}
    \label{theorem: LUCB correctness}
    The LUCB version for the representative identification problem stated in Algorithm \ref{algo: lucb} solves the problem with at least a probability of $1-\delta$
\end{theorem}

\textbf{Proof of Theorem \ref{theorem: LUCB correctness}:} 

The proof of Theorem \ref{theorem: LUCB correctness} follows a similar approach as that of Claim \ref{claim: vanilla correctness} with modified confidence bounds and an additional summation over $R_j^i$ going from $1$ to $R$. Basically on taking the union bound and using the inequality stated in \ref{hoeffding}, we have
\begin{align*}
    P[\text{Bad Event}] \leq \sum_{R=1}^\infty \sum_ {R_j^i =1}^R \sum_{i=1}^{m} \sum_{j = 1}^{c_i} 2 \text{exp}\Biggl(-2R_j^i \Biggl( \sqrt{\dfrac{\ln (\pi^2R^3N/3\delta)}{2R_j^i}} \Biggr)^2 \Biggr) 
    \leq \delta
\end{align*}

\end{document}